\documentclass[runningheads]{llncs}
\usepackage[T1]{fontenc}
\usepackage{graphicx}
\usepackage{amssymb}

\usepackage{algorithm}
\usepackage{algpseudocode}

\newcommand\blfootnote[1]{%
  \begingroup
  \renewcommand\thefootnote{}\footnote{#1}%
  \addtocounter{footnote}{-1}%
  \endgroup
}

\begin{document}
\title{ACC-UNet: A Completely Convolutional UNet model for the 2020s}
%
%
\author{Nabil Ibtehaz\inst{1}\orcidID{0000-0003-3625-5972} \and
Daisuke Kihara \inst{1,2,3}\orcidID{0000-0003-4091-6614} }

\authorrunning{N. Ibtehaz and D. Kihara}
%
%
%

\institute{Department of Computer Science, Purdue University, West Lafayette, IN, USA \and
Department of Biological Sciences, Purdue University, West Lafayette, IN, USA
\and Corresponding Author, \email{dkihara@purdue.edu}
}

\maketitle              
\begin{abstract}
This decade is marked by the introduction of Vision Transformer, a radical paradigm shift in broad computer vision. A similar trend is followed in medical imaging, UNet, one of the most influential architectures, has been redesigned with transformers. Recently, the efficacy of convolutional models in vision is being reinvestigated by seminal works such as ConvNext, which elevates a ResNet to Swin Transformer level. Deriving inspiration from this, we aim to improve a purely convolutional UNet model so that it can be on par with the transformer-based models, e.g, Swin-Unet or UCTransNet. We examined several advantages of the transformer-based UNet models, primarily long-range dependencies and cross-level skip connections. We attempted to emulate them through convolution operations and thus propose, ACC-UNet, a completely convolutional UNet model that brings the best of both worlds, the inherent inductive biases of convnets with the design decisions of transformers. ACC-UNet was evaluated on 5 different medical image segmentation benchmarks and consistently outperformed convnets, transformers, and their hybrids. Notably, ACC-UNet outperforms state-of-the-art models Swin-Unet and UCTransNet by  $2.64 \pm 2.54\%$ and $0.45 \pm 1.61\%$ in terms of dice score, respectively, while using a fraction of their parameters ($59.26\%$ and $24.24\%$). Our codes are available at \url{https://github.com/kiharalab/ACC-UNet}\blfootnote{Accepted at MICCAI 2023 Conference}.

\end{abstract}

\section{Introduction}

Semantic segmentation, an essential component of computer-aided medical image analysis, identifies and highlights regions of interest in various diagnosis tasks. However, this often becomes complicated due to various factors involving image modality and acquisition along with pathological and biological variations \cite{Olabarriaga2001InteractionSurvey}. The application of deep learning in this domain has thus certainly benefited in this regard. Most notably, ever since its introduction, the UNet model \cite{RonnebergerOlafandFischer2015U-Net:Segmentation} has demonstrated astounding efficacy in medical image segmentation. As a result, UNet and its derivatives have become the de-facto standard \cite{Zhou2021APromises}.

The original UNet model comprises a symmetric encoder-decoder architecture (Fig. 1a) and employs skip-connections, which provide the decoder spatial information probably lost during the pooling operations in the encoder. Although this information propagation through simple concatenation improves the performance, there exists a likely semantic gap between the encoder-decoder feature maps. This led to the development of a second class of UNets (Fig. 1b). U-Net++ \cite{Zhou2018Unet++:Segmentation} leveraged dense connections and MultiResUNet \cite{Ibtehaz2020MultiResUNet:Segmentation} added additional convolutional blocks along the skip connection as a potential remedy. 

Till this point in the history of UNet, all the innovations were performed using CNNs. However, the decade of 2020 brought radical changes in the computer vision landscape. The long-standing dominance of CNNs in vision was disrupted by vision transformers \cite{Dosovitskiy2021AnScale}. Swin Transformers \cite{Liu2021SwinWindows} further adapted transformers for general vision applications. Thus, UNet models started adopting transformers \cite{Chen2021TransUNet:Segmentation}. Swin-Unet \cite{CaoHuandWang2023Swin-Unet:Segmentation} replaced the convolutional blocks with Swin Transformer blocks and thus initiated a new class of models (Fig. 1c). Nevertheless, CNNs still having various merits in image segmentation, led to the development of fusing those two \cite{LinAiliangandXu2022ConTrans:Segmentation}. This hybrid class of UNet models (Fig. 1d) employs convolutional blocks in the encoder-decoder and uses transformer layers along the skip connections. UCTransNet \cite{Wang2022UCTransNet:Transformer} and MCTrans\cite{JiYuanfengandZhang2021Multi-compoundSegmentation} are two representative models of this class. Finally, there have also been attempts to develop all-transformer UNet architectures (Fig. 1e), for instance, SMESwin Unet \cite{WangZihengandMin2022SMESwinSegmentation} uses transformer both in encoder-decoder blocks and the skip-connection.

Very recently, studies have begun rediscovering the potential of CNNs in light of the advancements brought by transformers. The pioneering work in this regard is `A ConvNet for the 2020s' \cite{Liu2022A2020s}, which explores the various ideas introduced by transformers and their applicability in convolutional networks. By gradually incorporating ideas from training protocol and micro-macro design choices, this work enabled ResNet models to outperform Swin Transformer models.

\begin{figure}[t]
    \centering
    \includegraphics[width=\textwidth]{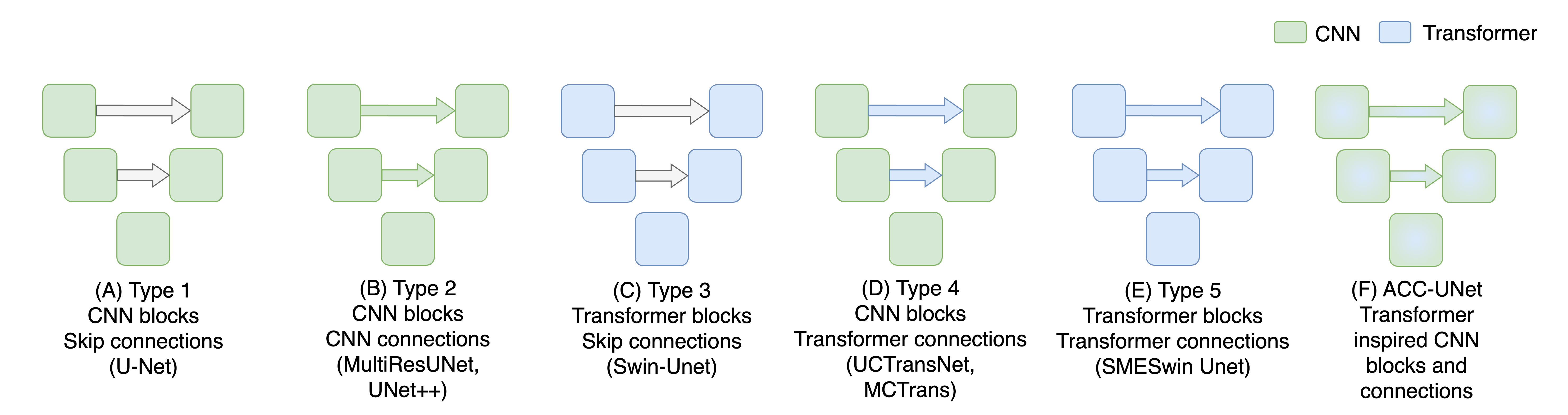}
    \caption{Developments and innovations in the UNet architecture.}
    \label{fig:types}
\end{figure}

In this paper, we ask the same question but in the context of UNet models. We investigate if a UNet model solely based on convolution can compete with the transformer-based UNets. In doing so, we derive motivations from the transformer architecture and develop a purely convolutional UNet model. We propose a patch-based context aggregation contrary to window-based self-attention. In addition, we innovate the skip connections by fusing the feature maps from multiple levels of encoders. Extensive experiments on 5 benchmark datasets suggest that our proposed modifications have the potential to improve UNet models.

\section{Method}

Firstly, we analyze the transformer-based UNet models from a high-level. Deriving motivation and insight from this, we design two convolutional blocks to simulate the operations performed in transformers. Finally, we integrate them in a vanilla UNet backbone and develop our proposed ACC-UNet architecture.

\begin{figure}
    \centering
    \includegraphics[width=\textwidth]{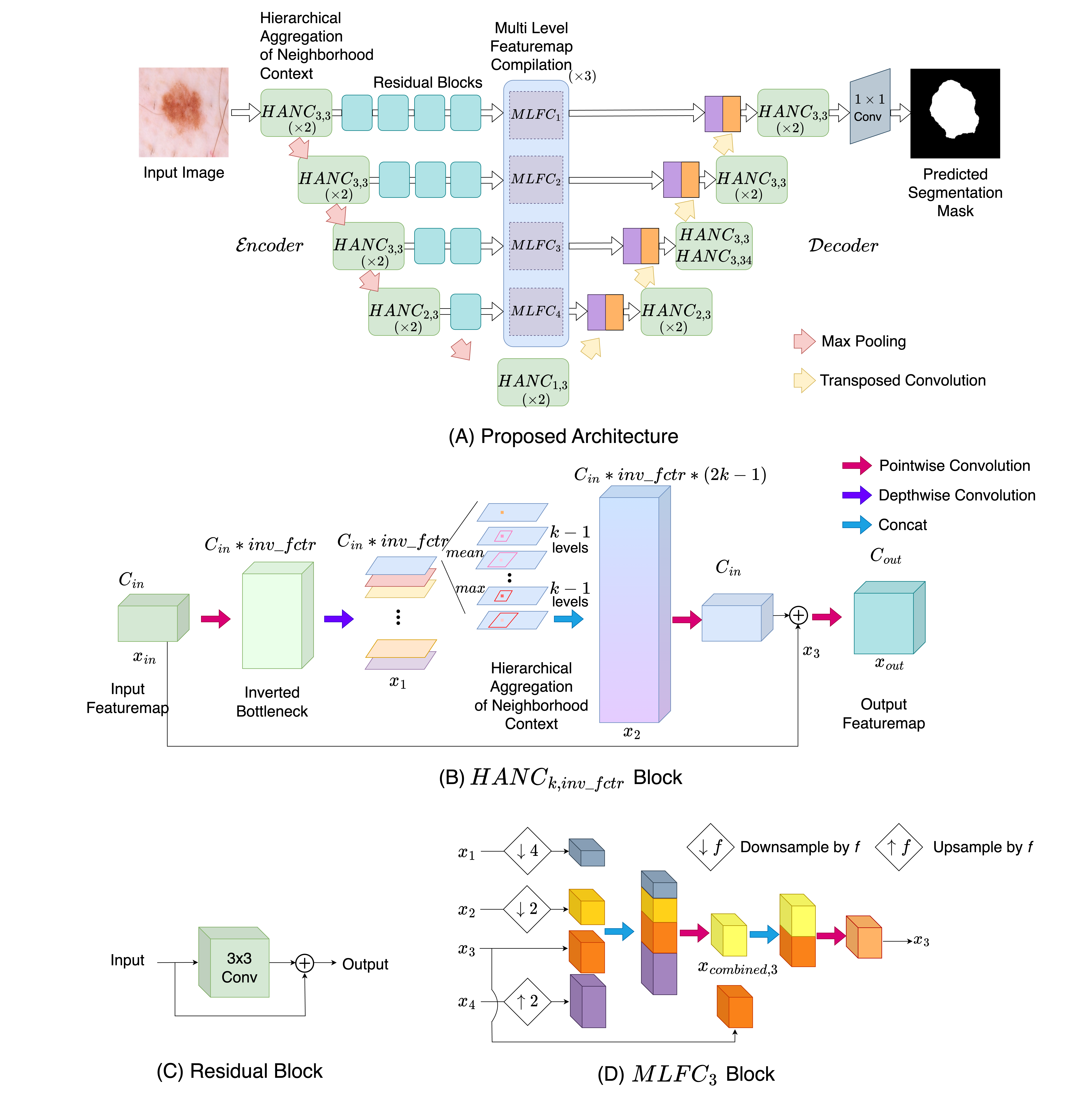}
    \caption{(A) Architecture of the proposed ACC-UNet. (B) A generalized view of $HANC_{k,inv\_fctr}$ block. (C) A generic residual block used in skip connection. (D) An example view of the 3rd level $MLFC$ block}
    \label{fig:fig2}
\end{figure}

\subsection{A high-level view of transformers in UNet}
Transformers apparently improve UNet models in two different aspects.

\subsubsection{Leveraging the long-range dependency of self-attention} 
Transformers can compute features from a much larger view of context through the use of (windowed) self-attention. In addition, they improve expressivity by adopting inverted bottlenecks, i.e., increasing the neurons in the MLP layer. Furthermore, they contain shortcut connections, which facilitate the learning \cite{Dosovitskiy2021AnScale}.

\subsubsection{Adaptive Multi-level feature combination through channel attention}
Transformer-based UNets fuse the feature maps from multiple encoder levels adaptively using channel attention. This generates enriched features due to the combination of various regions of interest from different levels compared to simple skip-connection which is limited by the information at the current level \cite{Wang2022UCTransNet:Transformer}.

Based on these observations, we modify the convolutional blocks and skip-connections in a vanilla UNet model to induce the capabilities of long-range dependency and multi-level feature combinations. 

\subsection{Hierarchical Aggregation of Neighborhood Context (HANC)}

We first explore the possibility of inducing long-range dependency along with improving expressivity in convolutional blocks. We only use pointwise and depthwise convolutions to reduce the computational complexity \cite{howard2017mobilenets}.

In order to increase the expressive capability, we propose to include inverted bottlenecks in convolutional blocks \cite{Liu2022A2020s}, which can be achieve by increasing the number of channels from $c_{in}$ to $c_{inv} = c_{in}*inv\_fctr$ using pointwise convolution. Since these additional channels will increase the model complexity, we use $3\times3$ depthwise convolution to compensate. An input feature map $x_{in} \in \mathbb{R}^{c_{in},n,m}$ is thus transformed to $x_1 \in \mathbb{R}^{c_{inv},n,m}$ as (Fig. 2b)

\begin{equation}
    x_1 = DConv_{3\times3}(PConv_{c_in \rightarrow c_{inv}}(x_{in}))
\end{equation}

Next, we wish to emulate self-attention in our convolution block, which at its core is comparing a pixel with the other pixels in its neighborhood \cite{Liu2021SwinWindows}. This comparison can be simplified by comparing a pixel value with the mean and maximum of its neighborhood. Therefore, we can provide an approximate notion of neighborhood comparison by appending the $mean$ and $max$ of the neighboring pixel features. Consecutive pointwise convolution can thus consider these and capture a contrasting view. Since hierarchical analysis is beneficial for images \cite{yang2022focal}, instead of computing this aggregation in a single large window, we compute this in multiple levels hierarchically, for example, $2\times2,2^2\times2^2,\cdots,2^{k-1}\times2^{k-1}$ patches. For $k=1$, it would be the ordinary convolution operation, but as we increase the value of $k$, more contextual information will be provided, bypassing the need for larger convolutional kernels. Thus, our proposed hierarchical neighborhood context aggregation enriches feature map $x_1 \in \mathbb{R}^{c_{inv},n,m}$ with contextual information as $x_2 \in \mathbb{R}^{c_{inv}*(2k-1),n,m}$ (Fig. 2b), where $||$ corresponds to concatenation along the channel dimension 
\begin{equation}
  \begin{array}{l}
    x_2 = (x_1 || mean_{2\times2}(x_1) || mean_{2^2\times2^2}(x_1) || \cdots || mean_{2^{k-1}\times2^{k-1}}(x_1)  \\ \;\;\;\;\;\;\;\;\;\;\;\;\;\;\;\;\;\;\;\; || max_{2\times2}(x_1) || max_{2^2\times2^2}(x_1) || \cdots || max_{2^{k-1}\times2^{k-1}}(x_1))
    \end{array}
\end{equation}

 Next, similar to the transformer, we include a shortcut connection in the convolution block for better gradient propagation. Hence, we perform another pointwise convolution to reduce the number of channels to $c_{in}$ and add with the input feature map. Thus, $x_2 \in \mathbb{R}^{c_{inv}*(2k-1),n,m}$ becomes $x_3 \in \mathbb{R}^{c_{in},n,m}$ (Fig. 2b)
 \begin{equation}
     x_3 = PConv_{c_{inv}*(2k-1) \rightarrow c_{in}}(x_2) + x_{in}
 \end{equation}
 
 Finally, we change the number of filters to $c_{out}$, as the output, using pointwise convolution (Fig. 2b)
 \begin{equation}
     x_{out} = PConv_{c_{in}\rightarrow c_{out}}(x_3)
 \end{equation}

 Thus, we propose a novel Hierarchical Aggregation of Neighborhood Context (HANC) block using convolution but bringing the benefits of transformers. The operation of this block is illustrated in Fig. 2b.
 
\subsection{Multi Level Feature Compilation (MLFC)}
Next, we investigate the feasibility of multi-level feature combination, which is the other advantage of using transformer-based UNets.

Transformer-based skip connections have demonstrated effective feature fusion of all the encoder levels and appropriate filtering from the compiled feature maps by the individual decoders \cite{JiYuanfengandZhang2021Multi-compoundSegmentation,Wang2022UCTransNet:Transformer,WangZihengandMin2022SMESwinSegmentation}. This is performed through concatenating the projected tokens from different levels \cite{Wang2022UCTransNet:Transformer}. Following this approach, we resize the convolutional feature maps obtained from the different encoder levels to make them equisized and concatenate them. This provides us with an overview of the feature maps across the different semantic levels. We apply pointwise convolution operation to summarize this representation and merge with the corresponding encoder feature map. This fusion of the overall and individual information is passed through another convolution, which we hypothesize enriches the current level feature with information from other level features. 

For the features, $x_1,x_2,x_3,x_4$ from 4 different levels, the feature maps can be enriched with multilevel information as (Fig. 2d)
\begin{equation}
    x_{comb,i} = PConv_{c_{tot}\rightarrow c_i}(resize_i(x1)||resize_i(x2)||resize_i(x3)||resize_i(x4))
\end{equation}
\begin{equation}
    x_i = PConv_{2c_i \rightarrow c_i}(x_{comb,i} || x_i), \;\;\;\;\;\;\;\;  i = 1,2,3,4
\end{equation}

Here, $resize_i(x_j)$ is an operation that resizes $x_j$ to the size of $x_i$ and $c_{tot}=c_1+c_2+c_3+c_4$. This operation is done individually for all the different levels.

We thus propose another novel block named Multi Level Feature Compilation (MLFC), which aggregates information from multiple encoder levels and enriches the individual encoder feature maps. This block is illustrated in Fig. 2d.

\subsection{ACC-UNet}
Therefore, we propose fully convolutional ACC-UNet (Fig. 2a). We started with a vanilla UNet model and reduced the number of filters by half. Then, we replaced the convolutional blocks from the encoder and decoder with our proposed HANC blocks. We considered $inv\_fctr = 3$, other than the last decoder block at level 3 ($inv\_fctr = 34$) to mimic the expansion at stage 3 of Swin Transformer. $k=3$, which considers up to $4\times4$ patches, was selected for all but the bottleneck level ($k=1$) and the one next to it ($k=2$). Next, we modified the skip connections by using residual blocks (Fig. 2c) to reduce semantic gap \cite{Ibtehaz2020MultiResUNet:Segmentation} and stacked 3 MLFC blocks. All the convolutional layers were batch-normalized \cite{Ioffe2015BatchShift}, activated by Leaky-RELU \cite{Maas2013RectifierModels} and recalibrated by squeeze and excitation \cite{Hu2018Squeeze-and-ExcitationNetworks}.

To summarize, in a UNet model, we replaced the classical convolutional blocks with our proposed HANC blocks that perform an approximate version of self-attention and modified the skip connection with MLFC blocks which consider the feature maps from different encoder levels.
The proposed model has $16.77$ M parameters, roughly a $2$M increase than the vanilla UNet model.

\section{Experiments}

\subsection{Datasets}
In order to evaluate ACC-UNet, we conducted experiments on 5 public datasets across different tasks and modalities. We used ISIC-2018 \cite{Codella2019SkinISIC,Tschandl2018TheLesions} (dermoscopy, 2594 images), BUSI \cite{Al-Dhabyani2020DatasetImages}(breast ultrasound, used 437 benign and 210 malignant images similar to \cite{ValanarasuJeyaMariaJoseandPatel2022UNeXt:Network}), CVC-ClinicDB \cite{Bernal2015WM-DOVAPhysicians} (colonoscopy, 612 images), COVID \cite{covid_db} (pneumonia lesion segmentation, 100 images), and GlaS \cite{Sirinukunwattana2017GlandContest} (gland segmentation, 85 training, and 80 test images). All the images and masks were resized to $224 \times 224$. For the GlaS dataset, we considered the original test split as the test data, for the other datasets we randomly selected $20\%$ of images as test data. The remaining $60\%$ and $20\%$ images were used for training and validation and the experiments were repeated 3 times with different random shuffling.

\subsection{Implementation Details}
We implemented ACC-UNet model in PyTorch and used a workstation equipped with AMD EPYC 7443P 24-Core CPU and NVIDIA RTX A6000 (48G) GPU for our experiments. We designed our training protocol identical to previous works  \cite{Wang2022UCTransNet:Transformer}, except for using a batch size of 12 throughout our experiments \cite{WangZihengandMin2022SMESwinSegmentation}. The models were trained for 1000 epochs \cite{WangZihengandMin2022SMESwinSegmentation} and we employed an early stopping patience of 100 epochs. We minimized the combined cross-entropy and dice loss \cite{Wang2022UCTransNet:Transformer} using the Adam \cite{Kingma2015} optimizer with an initial learning rate of $10^{-3}$, which was adjusted through cosine annealing learning rate scheduler \cite{ValanarasuJeyaMariaJoseandPatel2022UNeXt:Network} \footnote{Swin-UNet-based models were trained with SGD \cite{CaoHuandWang2023Swin-Unet:Segmentation} for poor performance of Adam}. We performed online data augmentations in the form of random flipping and rotating \cite{Wang2022UCTransNet:Transformer}.

\subsection{Comparisons with State-of-the-Art Methods}

\begin{table}[t]
\caption{Comparison with the state-of-the-art models. The first and second best scores are styled as bold and italic, respectively. The subscripts denote the standard deviation.}
\centering
\begin{tabular}{|c|c|c|c|c|c|c|c|}
\hline
Model        & params & FLOPs & ISIC-18                 & ClinicDB     & BUSI             & COVID            & GlaS                      \\ \hline
UNet        & 14M  & 37G   & $87.97_{0.11}$           & $90.66_{0.92}$  & $72.27_{0.86}$ & $71.21_{1.4}$  & $87.99_{1.32}$          \\ \hline
MultiResUNet & 7.3 M  & 1.1G    & $88.55_{0.24}$            & $88.20_{1.67}$  & $72.43_{0.91}$ & $71.33_{3.59}$ & $\mathit{88.34_{1.05}}$ \\ \hline
Swin-Unet    & 27.2 M  & 6.2G   & $\mathit{89.24_{0.14}}$ & $90.69_{0.50}$ & $76.06_{0.43}$ & $68.56_{1.07}$ & $86.45_{0.28}$          \\ \hline
UCTransnet & 66.4 M & 38.8G & $89.08_{0.44}$         & $\mathit{92.57_{0.39}}$ & $\mathit{76.56_{0.2}}$  & $\mathit{73.09_{3.63}}$ & $87.17_{0.85}$          \\ \hline
SMESwin-Unet & 169.8 M  & 6.4G   & $88.57_{0.13}$           & $89.62_{0.08}$  & $ 73.94_{2.06} $  & $58.4_{0.03}$  & $83.72_{0.18}$           \\ \hline
ACC-UNet   & 16.8 M & 38G & $\mathbf{89.37_{0.34}}$ & $\mathbf{92.67_{0.57}}$ & $\mathbf{77.19_{0.87}}$ & $\mathbf{73.99_{0.53}}$ & $\mathbf{88.61_{0.61}}$ \\ \hline
\end{tabular}
\end{table}

We evaluated ACC-UNet against UNet, MultiResUNet, Swin-Unet, UCTransnet, SMESwin-Unet, i.e., one representative model from the 5 classes of UNet, respectively (Fig. 1). Table 1 presents the dice score obtained on the test sets. 

The results show an interesting pattern. Apparently, for the comparatively larger datasets (ISIC-18) transformer-based Swin-Unet was the 2nd best method, as transformers require more data for proper training \cite{LinAiliangandXu2022ConTrans:Segmentation}. On the other end of the spectrum, lightweight convolutional model (MultiResUNet) achieved the 2nd best score for small datasets (GlaS). For the remaining datasets, hybrid model (UCTransnet) seemed to perform as the $2^{nd}$ best method. SMESwin-Unet fell behind in all the cases, despite having such a large number of parameters, which in turn probably makes it difficult to be trained on small-scale datasets. 

However, our model combining the design principles of transformers with the inductive bias of CNNs seemed to perform best in all the different categories with much lower parameters. Compared to much larger state-of-the-art models, for the 5 datasets, we achieved $0.13\%,0.10\%,0.63\%,0.90\%, 0.27\%$ improvements in dice score, respectively. Thus, our model is not only accurate, but it is also efficient in using the moderately small parameters it possesses. In terms of FLOPs, our model is comparable with convolutional UNets, the transformer-based UNets have smaller FLOPs due to the massive downsampling at patch partitioning.

\begin{figure}[h]
    \centering
    \includegraphics[width=\textwidth]{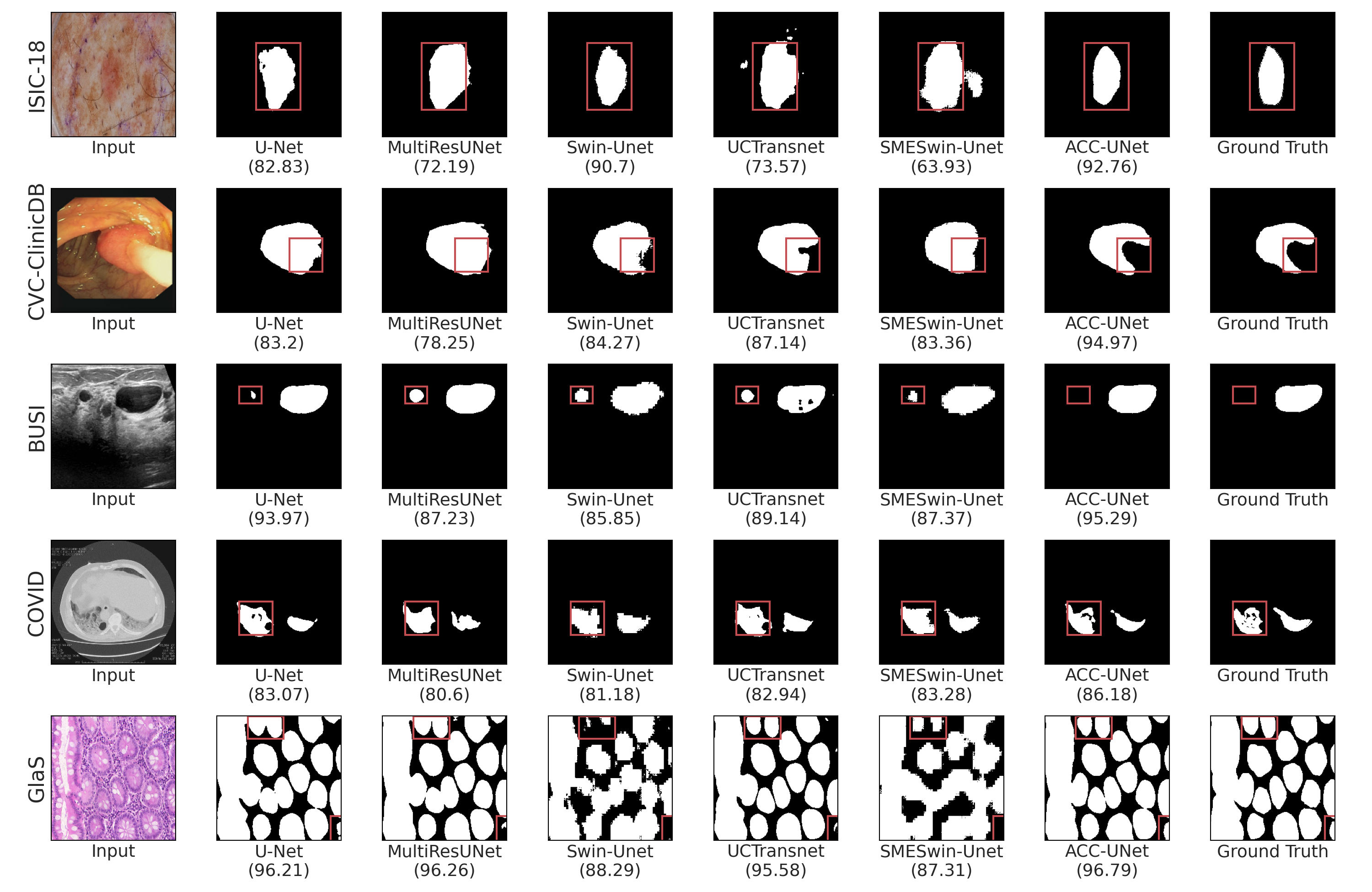}
    \caption{Comparative qualitative results, with dice score provided inside the parenthesis.}
    \label{fig:fig3}
\end{figure}

\subsection{Comparative Qualitative Results on the Five Datasets}
In addition to, achieving higher dice scores, apparently, ACC-UNet generated better qualitative results. Fig. 3 presents a qualitative comparison of ACC-UNet with the other models. Each row of the figure comprises one example from each of the datasets and the segmentation predicted by ACC-UNet and the ground truth mask are presented in the rightmost two columns. For the $1^{st}$ example from the ISIC-18 dataset, our model did not oversegment but rather followed the lesion boundary. In the $2^{nd}$ example from CVC-ClinicDB, our model managed to distinguish the finger from the polyp almost perfectly. Next in the $3^{rd}$ example from BUSI, our prediction filtered out the apparent nodule region on the left, which was predicted as a false positive tumor by all the other models. Similarly, in the $4^{th}$ sample from the COVID dataset, we were capable to model the gaps in the consolidation of the left lung visually better, which in turn resulted in $2.9\%$ higher dice score than the $2^{nd}$ best method. Again, in the final example from the GlaS dataset, we not only successfully predicted the gland at the bottom right corner but also identified the glands at the top left individually, which were mostly missed or merged by the other models, respectively.

\subsection{Ablation Study}

\begin{figure}
    \centering
\includegraphics[width=0.8\textwidth]{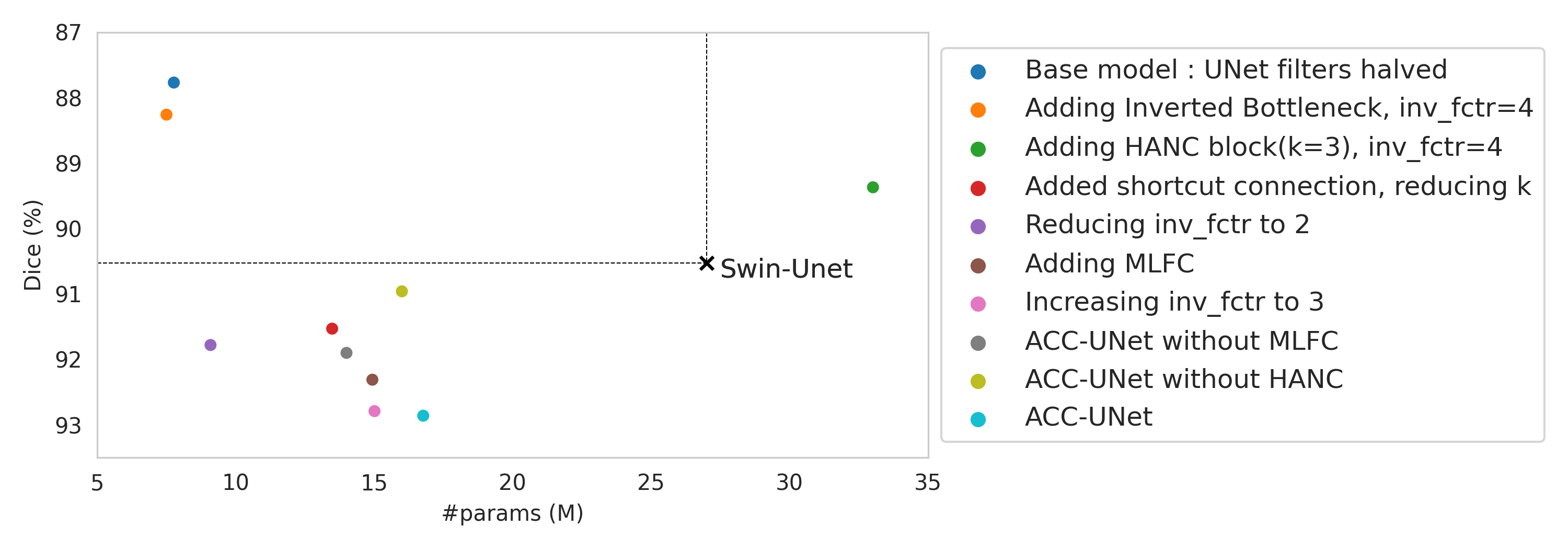}
    \caption{Ablation study on the CVC-ClinicDB dataset.}
    \label{fig:fig4}
\end{figure}

We performed an ablation study on the CVC-ClinicDB dataset to analyze the contributions of the different design choices in our roadmap (Fig. \ref{fig:fig4}). We started with a UNet model with the number of filters halved as our base model, which results in a dice score of $87.77\%$ with $7.8M$ parameters. Using depthwise convolutional along with increasing the bottleneck by 4 raised the dice score to $88.26\%$ while slightly reducing the parameters to $7.5M$. Next, HANC block was added with $k=3$ throughout, which increased the number of parameters by $340\%$ for an increase of $1.1\%$ dice score. Shortcut connections increased the performance by $2.16\%$. We also slowly reduced both $k$ and $inv\_fctr$ which reduced the number of parameters without any fall in performance. Finally, we added the MLFC blocks (4 stacks) and gradually optimized $k$ and $inv\_fctr$ along with dropping one MLFC stage, which led to the development of ACC-UNet. Some other interesting ablations were ACC-UNet without MLFC (dice $91.9\%$) or HANC (dice $90.96\%$, with $25\%$ more filters to keep the number of parameters comparable).

\section{Conclusions}
Acknowledging the benefits of various design paradigms in transformers, we investigate the suitability of similar ideas in convolutional UNets. The resultant ACC-UNet possesses the inductive bias of CNNs infused with long-range and multi-level feature accumulation of transformers. Our experiments reveals this amalgamation indeed has the potential to improve UNet models. One limitation of our model is the slowdown from concat operations (please see supplementary materials), which can be solved by replacing them. In addition, there are more innovations brought by transformers \cite{Liu2022A2020s}, e.g., layer normalization, GELU activation, AdamW optimizer, these will be explored further in our future work.


\subsubsection{Acknowledgements}This work was partly supported by the National Institutes of Health (R01GM133840 and 3R01GM133840-02S1) and the National Science Foundation (CMMI1825941, MCB1925643, IIS2211598, DMS2151678, DBI2146026, and DBI2003635).

\bibliographystyle{splncs04}
\bibliography{main}

\clearpage

\appendix

\section{Remarks on Training Time}

In this work, most of our focus was put on bringing attributes from transformers to CNNs. In achieving this, we relied on straightforward implementation, and so did not optimize our architecture thoroughly. As a result, our model is comparatively slow. We have presented the training and inference time of the models on the BUSI dataset in the following table.

\begin{table}[h]
\caption{Training and Inference Time on the BUSI Dataset}
\centering
\begin{tabular}{|c|c|c|}
\hline
\textbf{Model}       & \textbf{Training sec/step} & \textbf{Inference sec/step} \\ \hline
UNet                 & 0.46                       & 0.39                        \\ \hline
MultiresUNet         & 0.75                       & 0.53                        \\ \hline
Swin-Unet            & 0.55                       & 0.39                        \\ \hline
UCTransNet           & 1.33                       & 0.7                         \\ \hline
SMESwin-Unet         & 2.6                        & 1.89                        \\ \hline
ACC-UNet             & 2.11                       & 0.61                        \\ \hline
ACC-UNet (no concat) & 1.51                       & 0.53                        \\ \hline
\end{tabular}
\end{table}

This slowdown primarily comes from the concat operation, which is extremely slow in the current pytorch implementation. However, we can avoid most of them by using a pre-allocated tensor and populating it with the computed featuremaps. This reduces the training time by $28.43\%$ and makes it comparable to UCTransNet. The inference time is also benefitted from this. Further optimization is possible to reduce the computational time, which we will focus on in our immediate future work. 

\clearpage

\section{Additional Qualitative Results}
\begin{figure}[h]
    \centering
\includegraphics[width=0.8\textwidth]{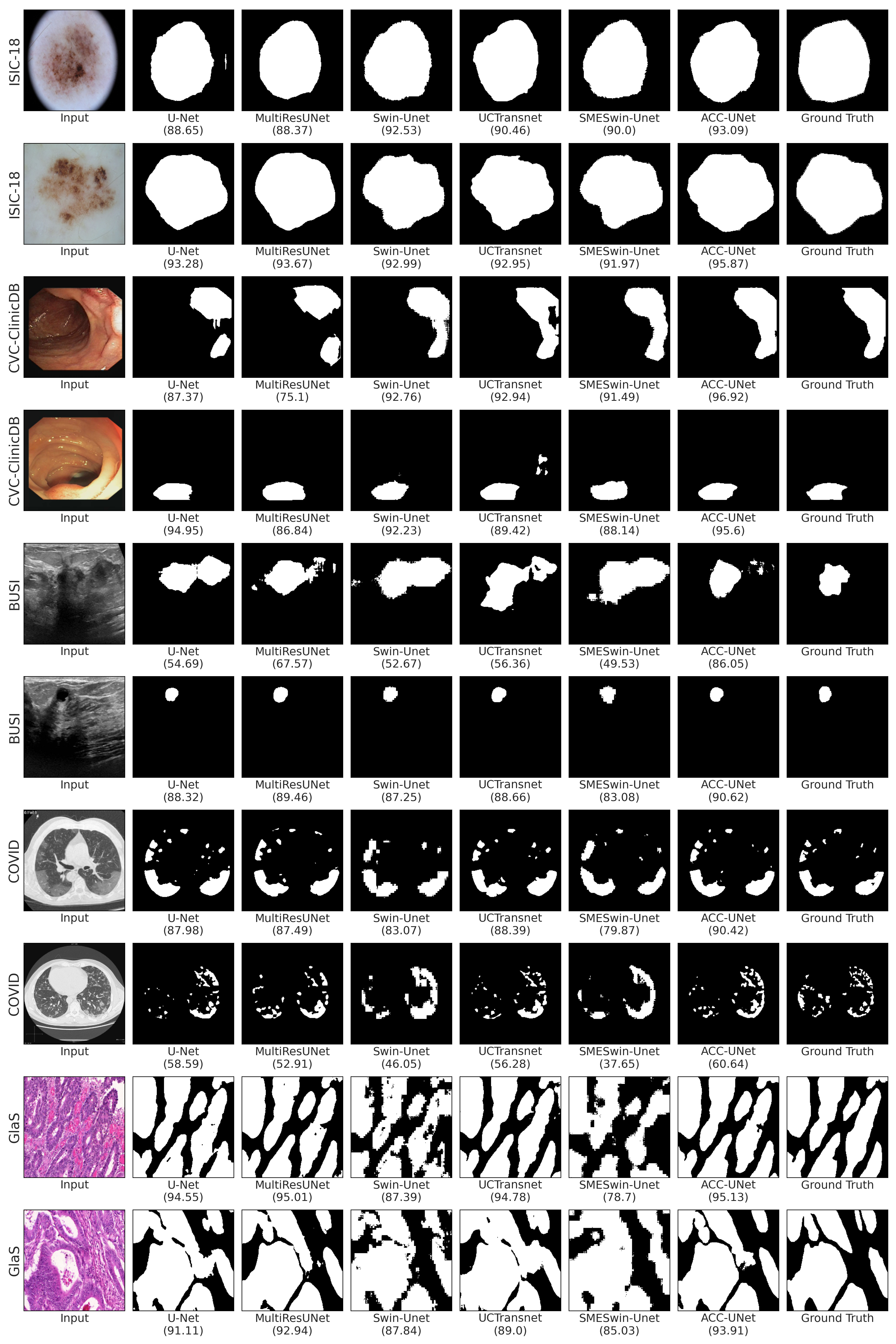}
\end{figure}

\end{document}